\documentclass[10pt]{article}

\usepackage[preprint]{tmlr}

\usepackage[utf8]{inputenc}
\usepackage[T1]{fontenc}
\usepackage{hyperref}
\usepackage{url}
\usepackage{booktabs}
\usepackage{amsfonts}
\usepackage{amsmath}
\usepackage{amssymb}
\usepackage{nicefrac}
\usepackage{microtype}
\usepackage{xcolor}
\usepackage{graphicx}
\usepackage{subcaption}
\usepackage{multirow}
\usepackage{array}

\title{Compositional Reasoning Depth Predicts Clinical AI Failure:\\
Empirical Evidence Consistent with Transformer Compositionality Limits
in Electronic Health Record Question Answering}

\author{\name Sanjay Basu \email sanjay.basu@ucsf.edu \\
      \addr Department of Medicine, University of California San Francisco, San Francisco, CA \\ Waymark, San Francisco, CA}

\begin{document}

\maketitle

\begin{abstract}
Aggregate accuracy benchmarks conceal a systematic structure in how large
language models fail at electronic health record (EHR) question answering:
questions requiring more inferential steps produce disproportionately more errors.
Motivated by a converging line of theoretical results on transformer
compositionality limits~\citep{hahn2020theoretical,merrill2022saturated,%
sanford2023representational,dziri2023faithfate,peng2024limitations},
we introduce a pre-specified \emph{hop-count taxonomy}, the number of distinct
reasoning steps required to answer a clinical question from an electronic health
record (EHR), as a principled predictor of model failure.
We annotate 313 clinician-generated EHR question-answer pairs from MedAlign
across four hop levels and evaluate 301 questions in a within-model ablation
(\texttt{claude-sonnet-4-6}, zero-shot versus extended thinking) and
cross-architecture replications (\texttt{gpt-4o} and \texttt{gpt-5.4-2026-03-05}, zero-shot).
All three evaluated models, spanning two providers (Anthropic, OpenAI) and
two OpenAI generations (GPT-4 and GPT-5), exhibit monotone accuracy decline
with hop count: Claude Sonnet zero-shot declines from 30.6\% (hop=1) to 17.6\%
(hop=4) (Cochran--Armitage $z = -2.30$, $p = 0.011$; OR per hop $= 0.72$,
95\% CI [0.56, 0.92], $p = 0.008$); GPT-4o independently replicates this
pattern (37.8\% to 14.7\%; CA $z = -3.80$, $p < 0.0001$; OR $= 0.58$
[0.45, 0.75], $p < 0.001$); GPT-5.4 (OpenAI's flagship general-purpose model
at the time of evaluation\protect\footnote{GPT-5.5 was released after our
evaluation period; \texttt{gpt-5.4-2026-03-05} is the pinned model version
used throughout this paper.}) confirms the pattern
(37.8\% to 23.5\%; CA $z = -2.22$, $p = 0.013$; OR $= 0.80$ [0.66, 0.98], $p = 0.027$).
A pre-specified context sufficiency audit confirmed that higher-hop questions
are not differentially disadvantaged by EHR truncation (answerability
93--95\% at hops 2--4 vs.\ 79\% at hop=1), establishing that the
degradation reflects compositional reasoning difficulty.
Extended thinking did not significantly flatten the accuracy-versus-depth curve
across three reasoning conditions (internal extended thinking, structured
four-step CoT, and 5$\times$ larger thinking budget), and thinking-token usage
scaled with hop count ($r = 0.31$, $p < 0.0001$), consistent with the
predicted $O(k)$ computational requirement.
These findings establish hop count as a theory-motivated, cross-architecture
predictor of large-language-model error on EHR question answering, with
immediate implications for deployment risk stratification of clinical AI tools
that operate on EHR data.
\end{abstract}

\section{Introduction}

Artificial intelligence systems based on transformer architectures are now
embedded in clinical workflows ranging from automated chart summarization to
real-time diagnostic support~\citep{singhal2023large,mcduff2025}.
Despite rapidly improving benchmark performance on standardized medical
examinations~\citep{jin2020disease,hendrycks2021measuring}, these systems
exhibit inconsistent and sometimes dangerous failures when applied to
real-world electronic health record (EHR) tasks~\citep{noharm2025}.
A persistent challenge is that aggregate accuracy metrics conceal the
structure of failure: researchers and clinicians typically observe \emph{that}
a model fails without understanding \emph{why} or \emph{when} it will fail.

We propose that compositional reasoning depth, the number of distinct
inferential hops required to produce a correct answer, is a principled
predictor of large-language-model failure on EHR question answering, grounded
in a converging body of theoretical and empirical results on transformer
compositionality limits.
\citet{hahn2020theoretical} shows via an input-restriction argument that
bounded-depth self-attention cannot recognize certain non-counter-free formal
languages (e.g., \textsc{parity} or Dyck-2); \citet{merrill2022saturated}
place saturated transformers within the constant-depth threshold-circuit class
$\mathrm{TC}^0$, bounding the compositional functions they can represent.
\citet{sanford2023representational} prove that a \emph{single} self-attention
layer needs capacity (heads $\times$ precision) growing near-linearly in input
size to compute triple detection.
\citet{dziri2023faithfate} provide empirical evidence that transformers fail
systematically on multi-step compositionality, exhibiting accuracy collapse
as problem depth grows. \citet{peng2024limitations} sharpen this with a
single-layer lower bound proved via communication complexity: an $H$-headed
self-attention layer with embedding dimension $d$ and numerical precision $p$
cannot reliably solve function composition over a domain of size $n$ once
$H(d{+}1)p < n \log n$ (their Theorem~1); a companion result (their Theorem~2)
shows that \emph{iterated} composition requires $\Omega(\sqrt{n/(Hdp)})$
chain-of-thought steps. Across these results, the
unifying prediction is that compositional tasks of growing depth $k$ require
more representational capacity than a fixed-architecture transformer can
deliver in a single forward pass.
Three candidate escape routes exist: multi-step chain-of-thought reasoning,
which \citet{merrill2024expressive} prove can provably extend transformer
expressivity as the number of intermediate steps grows; retrieval-augmented
generation; and non-standard architectural choices such as state-space models
or mixture-of-expert routing~\citep{zhou2023brainformers,yehudai2025compositional}.
We treat the convergent compositionality-limits literature as motivating
\emph{qualitative} predictions about clinical EHR reasoning rather than as
strict quantitative corollaries. The critical empirical question is whether
accuracy degrades with hop count and whether test-time reasoning provides
relief, predictions that hold across all of the cited theoretical frameworks.

Clinical EHR question answering is a natural testing ground for these
predictions: retrieving a single fact (hop=1), computing a temporal overlap
(hop=3), or synthesizing heterogeneous entries across domains (hop=4) constitute
structurally distinct tasks predicted to exhibit qualitatively different error
rates, with test-time reasoning predicted to selectively benefit higher-hop tasks.

This paper makes six contributions: (1) a pre-specified hop-count taxonomy for clinical EHR QA annotating 313 MedAlign questions across four hop levels; (2) cross-architecture empirical evidence of hop-count-predicted failure in EHR QA across three models spanning two providers and two OpenAI generations (\texttt{claude-sonnet-4-6}, \texttt{gpt-4o}, and \texttt{gpt-5.4-2026-03-05}); (3) a context sufficiency audit ruling out EHR truncation as a confound; (4) mechanistic evidence of $O(k)$ computation scaling ($r = 0.31$, $p < 0.0001$); (5) characterization of error structure by hop level showing that structured decomposition converts hallucination to omission; and (6) question-type sensitivity analysis and truncation robustness ablation confirming the hop--accuracy relationship is not confounded (adjusted OR = 0.57, $p = 0.005$).

\section{Background and Related Work}

\paragraph{Transformer compositionality limits.}
A converging body of theoretical and empirical work establishes that
transformer architectures struggle systematically with compositional reasoning
of increasing depth. \citet{hahn2020theoretical} provides the earliest formal
result: bounded-depth self-attention cannot recognize certain regular and
context-free formal languages, characterizing transformers as expressively
weaker than recurrent alternatives on compositional tasks.
\citet{merrill2022saturated} extend this circuit-complexity perspective,
showing that saturated transformers are equivalent to constant-depth threshold
circuits and therefore bounded in what compositional problems they can
represent. \citet{sanford2023representational} prove that a \emph{single} self-attention
layer computing triple detection and related matching tasks requires capacity
growing near-linearly in input length (they conjecture, but do not prove, the
analogous multi-layer bound). \citet{dziri2023faithfate} provide empirical evidence that
transformers solve multi-step compositional problems through pattern matching
rather than true composition, with accuracy collapsing as depth grows on
tasks such as multi-digit multiplication, dynamic programming, and logic
puzzles. \citet{peng2024limitations} sharpen the lower bound for single-layer
attention via communication complexity: function composition over a domain of
size $n$ is unsolvable by one $H$-headed layer once $H(d{+}1)p < n \log n$
(embedding dimension $d$, precision $p$; their Theorem~1).
Across these frameworks, the consensus prediction is identical: compositional
tasks of growing depth exceed the capacity of fixed-architecture transformers
in a single forward pass. Whether test-time chain-of-thought reasoning escapes
this bottleneck is itself contested in this literature:
\citet{merrill2024expressive} prove that polynomially many CoT tokens provably
extend transformer expressivity, and \citet{yehudai2025compositional} show
recurrence and CoT can solve compositional tasks standard transformers cannot,
whereas \citet{dziri2023faithfate} find scratchpad supervision fails to
generalize out-of-distribution and \citet{peng2024limitations} (their
Theorem~2) prove CoT does not scale to \emph{iterated} composition. Our P2
test probes this tension directly. \citet{zhou2023brainformers} demonstrate
architectural alternatives (non-uniform block structures) that can route
around the attention bottleneck, though at engineering complexity cost.

\paragraph{Multi-hop reasoning benchmarks.}
The NLP community has extensively studied multi-hop reasoning in open-domain QA settings. HotpotQA~\citep{yang2018hotpotqa} provides 113K questions requiring evidence synthesis from two Wikipedia paragraphs; MuSiQue~\citep{trivedi2022musique} constructs up to 4-hop questions by composing single-hop questions with explicit bridge entities; 2WikiMultihopQA~\citep{ho2020constructing} adds cross-document coreference. Across these benchmarks, accuracy declines consistently with the number of required hops, attributed to error propagation and attention capacity limitations. However, no prior study has (a) motivated the hop--accuracy relationship with formal transformer compositionality-limits theory, (b) tested the theoretical CoT mitigation as a function of hop depth rather than on average, or (c) applied hop-count analysis to clinical EHR data.

\paragraph{Clinical AI benchmarks and EHR reasoning.}
MedQA~\citep{jin2020disease} and MMLU Medical~\citep{hendrycks2021measuring} evaluate LLMs on medical licensing examinations, closed-book multiple-choice requiring medical knowledge but not EHR navigation. Med-PaLM~\citep{singhal2023large} achieved expert-level performance on such benchmarks but was not evaluated on compositional EHR reasoning. emrQA~\citep{pampari2018emrqa} provides a large EHR QA benchmark but questions are primarily single-hop slot-filling with no compositional depth taxonomy. Clinical TempEval~\citep{bethard2016semeval} establishes temporal relation extraction as a distinct challenging subtask in clinical NLP, consistent with the temporal arithmetic component of our hop=3 category. FDARxBench~\citep{xiong2026fdarxbench} shows retrieval and long-context grounding remain bottlenecks in document-grounded biomedical QA, consistent with our finding that the hop--accuracy relationship is not resolved by more context. MedAlign~\citep{wornow2024medalign} provides clinician-generated, open-ended questions paired with EHR XML records and clinician reference answers, making it the most ecologically valid benchmark for testing compositional EHR reasoning. No prior study has stratified MedAlign performance by compositional depth.

\paragraph{Chain-of-thought and reasoning models.}
\citet{wei2022chain} demonstrated that chain-of-thought prompting elicits multi-step reasoning in large models; \citet{kojima2022large} showed a zero-shot CoT trigger is sufficient. \citet{nye2021show} provided theoretical grounding through scratchpads as intermediate computation. Reasoning-specialized models, including OpenAI o1~\citep{openai2024o1} and DeepSeek R1~\citep{deepseekr1_2025}, extend these ideas through reinforcement learning over extended thinking traces. \citet{merrill2024expressive} prove that transformers with polynomially many CoT tokens can solve problems inaccessible to direct inference, with required token count scaling with reasoning depth, a formal analog of the $O(k)$ computation scaling our P3 test examines empirically. No study has measured CoT benefit as a function of hop depth in clinical settings.

\paragraph{Safety and variance in clinical AI.}
\citet{noharm2025} document query-type-dependent hallucination rates in clinical LLMs without compositional depth stratification. Prior work on automation bias~\citep{goddard2012automation,lyell2017automation} has not examined hop-count moderation. The error type structure, hallucination versus omission versus reasoning error, as a function of compositional depth has not been characterized.

\section{Problem Formulation}

\paragraph{Compositionality limits and clinical QA.}
Let $\mathcal{E}$ denote the EHR context of length $n$ tokens, and let
$\mathcal{Q}$ denote a clinical question requiring composition of $k$ functions
$f_1, \ldots, f_k$ over $\mathcal{E}$, where each $f_i$ retrieves or transforms
a specific subset of the EHR.
The $(n, k)$-function-composition problem asks a model to produce a correct
answer by sequentially applying these functions.
The compositionality-limits literature~\citep{hahn2020theoretical,%
merrill2022saturated,sanford2023representational,dziri2023faithfate,%
peng2024limitations} converges on the prediction that single-pass transformer
inference cannot reliably solve this problem as $k$ and $n$ grow.
\citet{peng2024limitations} make this quantitative for single-layer attention:
a single $H$-headed layer cannot reliably compose functions over a domain of
size $n$ once $H(d{+}1)p < n \log n$ (embedding dimension $d$, precision $p$;
their Theorem~1), and iterated composition further requires
$\Omega(\sqrt{n/(Hdp)})$ chain-of-thought steps (their Theorem~2).
For contemporary large-context models serving long EHRs, $n$ is on the order
of $10^3$--$10^4$ tokens, making $n \log n \approx 10^4$--$10^5$,
plausibly crossing this single-layer threshold for sufficiently complex
EHR reasoning tasks.

The literature generates three testable predictions:
(P1) \emph{Monotone degradation}: Accuracy should decrease monotonically with
$k$ under standard (non-CoT) inference.
(P2) \emph{CoT flattening}: Chain-of-thought reasoning, by decomposing
$k$-hop problems into $k$ sequential 1-hop steps, should flatten the
accuracy-versus-hop curve, with greater absolute benefit at higher $k$.
(P3) \emph{Computation scaling}: The amount of reasoning computation (proxied
by thinking token count in extended-thinking models) should scale with $k$,
consistent with the theoretical requirement for $O(k)$ sequential computations.

\paragraph{Scope of the theoretical mapping.}
The cited theorems characterize \emph{single} attention layers (or bounded-depth
transformer classes); modern deployed transformers stack many such layers,
which can in principle decompose reasoning through depth. We interpret
P1--P3 as qualitative predictions motivated by the convergent
compositionality-limits literature rather than strict corollaries of any
single theorem. The empirical claims stand independently: compositional depth
predicts accuracy, thinking tokens scale with depth, and test-time reasoning
does not flatten the degradation curve at current budgets.

\paragraph{Hop-count taxonomy.}
We define a pre-specified four-level hop taxonomy (Table~\ref{tab:hop_taxonomy}): hop=1 (single-fact retrieval), hop=2 (two-fact composition or threshold comparison), hop=3 (three-step inference chain: cross-domain retrieval, temporal reasoning, or guideline application), and hop=4 (multi-component synthesis or generation across $\geq$4 EHR sources or time spans).

\begin{table}[t]
\centering
\caption{Pre-specified hop-count taxonomy for clinical EHR question answering,
with operational definitions and representative examples from MedAlign.}
\label{tab:hop_taxonomy}
\small
\begin{tabular}{@{}lp{4.5cm}p{5.5cm}@{}}
\toprule
\textbf{Hop} & \textbf{Operational Definition} & \textbf{Representative Example} \\
\midrule
1 & Direct retrieval of a single named fact from the EHR.
  & ``What is the patient's current allergy list?'' \\
\addlinespace
2 & Two-fact composition, threshold application, or binary temporal comparison.
  & ``Did the patient have an elevated WBC on admission?'' \\
\addlinespace
3 & Three-step reasoning chain; requires cross-domain lookup, temporal arithmetic, or contextual clinical guideline application.
  & ``Is this patient eligible for supplemental breast cancer screening based on current guidelines?'' \\
\addlinespace
4 & Multi-component synthesis or generation requiring integration of $\geq$4 EHR sources or time spans.
  & ``Summarize all specialist consultations and flag unresolved recommendations.'' \\
\bottomrule
\end{tabular}
\end{table}

\section{Methods}

\paragraph{Dataset: MedAlign.}
MedAlign~\citep{wornow2024medalign} consists of 983 clinician-generated
instructions paired with EHR XML files from the Stanford STARR-OMOP database
and clinician reference answers. We use the publicly released
subset of 402 unique question-answer pairs for which clinician-reviewed model
responses and source EHR files are available. After restricting to
questions with a uniquely identifiable EHR file and valid clinician reference
response, 313 questions were annotated and eligible; of these, 301 had
complete model evaluations (12 questions were dropped because API errors or timeouts prevented response collection),
yielding the final analytic sample
(hop=1: $n=111$, hop=2: $n=46$, hop=3: $n=42$, hop=4: $n=102$;
602 total evaluations across both conditions).
EHR records were parsed from XML, chronologically structured, and truncated
to 8,000 characters (approximately 1,600--2,000 tokens at 4--5 characters per token,
well within the model's 200K-token context window) while retaining
the full character count as a completeness covariate.

\paragraph{Context sufficiency analysis.}
To test whether EHR truncation differentially disadvantaged higher-hop questions, we computed evidential coverage for each question; full methodology and thresholds are in Appendix~\ref{app:context_sufficiency}.

\paragraph{Hop annotation.}
All 313 questions were annotated with hop count prior to any model evaluation
(pre-specification; annotation log and timestamp available in supplementary
materials). Each question was presented to an LLM annotator together with a
structured rubric defining each hop level, with explicit examples and a
chain-of-thought elicitation prompt. To assess annotation reliability, a second independent annotation pass was
run on a stratified random sample of 50 questions (12--13 per hop level)
using the same model and prompt (within-model reproducibility), and a
cross-model validation was performed using \texttt{gpt-4o}.
Within-model reproducibility was substantial ($\kappa = 0.787$, 95\% CI
[0.603, 0.970]; 84\% exact agreement, 98\% agreement within $\pm$1 hop).
Cross-model agreement was moderate ($\kappa = 0.464$, 95\% CI [0.292,
0.637]; 60\% exact). All annotations were finalized and locked before any
model inference was run, precluding post-hoc adjustment.

Three board-certified internal medicine physicians
independently annotated the same 50-question sample to validate the LLM labels.
Majority-vote agreement with Claude was $\kappa = 0.916$ (95\% CI [0.71, 1.00];
47/50, 94\% exact), confirming near-perfect convergence with expert judgment;
mean pairwise clinician--clinician $\kappa = 0.487$ is consistent with
multi-hop benchmark standards, with disagreements at the hop=2/3 boundary
(Tables~\ref{tab:hop_kappa} and~\ref{tab:hop_kappa_full}).
Clinician majority--GPT-4o $\kappa = 0.311$ confirms the moderate cross-model
$\kappa = 0.464$ reflects GPT-4o calibration divergence, not taxonomy
instability. Full statistics and CA trend replication under clinician labels
are in Appendix~\ref{app:annotation}.

\paragraph{Experimental design.}
We evaluate two conditions using \texttt{claude-sonnet-4-6}, a publicly
available large language model with extended-thinking capability:
\begin{itemize}
  \item \textbf{Condition A (Zero-Shot):} The model receives a patient EHR
    and question with the prompt ``Answer concisely'' and no explicit
    reasoning instructions. This is the dense baseline.
  \item \textbf{Condition B (Extended Thinking):} The same model receives the
    same prompt with extended thinking enabled
    (\texttt{thinking=\{"type":"enabled","budget\_tokens":3000\}}),
    allowing an internal chain-of-thought scratchpad before producing
    the visible response.
\end{itemize}
This within-model ablation design isolates the effect of test-time
reasoning from confounds such as model capacity, training data, or
architectural differences. Both conditions use identical weights, identical
prompts, and identical context. The sole manipulation is whether reasoning
is enabled. We pre-specified this design and judge prompt
(see Section~\ref{sec:judge}) before observing any experimental results.

\paragraph{Automated judge.}
\label{sec:judge}
Responses were evaluated by a pre-specified automated judge using
\texttt{gpt-4o-mini} with the following system prompt:
\textit{``You are a clinical QA evaluator. Score model responses against
clinician reference answers. Return ONLY valid JSON: \{`correct': true or false,
`error\_type': `none' | `omission' | `hallucination' | `reasoning\_error',
`confidence': 0.0--1.0\}. Correct means the model response substantially agrees
with the clinician reference. Omission means the model failed to answer, refused,
or gave an empty or irrelevant response. Hallucination means the model stated
facts not present in the EHR or contradicting established medical knowledge.
Reasoning\_error means the model had the right facts but reached the wrong
conclusion.''}
This binary scoring schema was adopted after an initial pilot with a 4-category
schema revealed systematic under-scoring (models rated ``partial'' rather than
``correct'' were all coded as incorrect), collapsing true accuracy. Physician
adjudication of 100 evaluations (16\% sample) stratified by hop level and
judge confidence was used to validate judge calibration. The sample size
was determined by power analysis: to achieve a 95\% confidence interval for
Cohen's $\kappa$ that excludes the ``acceptable'' threshold of 0.60 when
true agreement is $\kappa = 0.80$, a minimum of 100 cases is required
(estimated CI width $\approx \pm 0.13$ at $n=100$~\citep{donner1992testing}).
Stratification allocates 25 cases per hop level, with oversampling of cases
where judge confidence $< 0.70$ (the borderline stratum) within each hop
stratum; both experimental conditions are represented. Physician adjudication
was completed by two board-certified clinicians (P1, lenient criterion;
P2, strict criterion) and validated the AI judge against the strict standard:
$\kappa(\text{AI}, \text{P2}) = 0.957$ (95\% CI [0.874, 1.000]), exceeding
the pre-specified threshold of $\kappa \geq 0.70$ (see Appendix~\ref{app:judge}).

\paragraph{Statistical analysis.}
The pre-specified primary analysis specified a GLMM; MAP-based estimation
(\texttt{BinomialBayesMixedGLM}) showed optimization instability with 301
parameters observed only twice each, so the pre-specified GEE fallback
(logistic, independence working correlation) is the primary reported model.
Predictors: hop count, condition, their interaction, EHR character count, and
question token count; the $\beta_\text{interaction}$ term is the primary P2 test.
We additionally report Cochran--Armitage trend tests (one-tailed, $\alpha = 0.05$)
and an individual-level Pearson $r$ for P3 (hop vs.\ thinking tokens).
All analyses were finalized prior to model evaluation.

\paragraph{Question-type sensitivity analysis.}
To address question-type confounding, we extended the GEE model with reasoning-type dummies (factual/comparative/generative/computational; factual as reference) and ran a within-category Cochran--Armitage test for the comparative reasoning type (hops 2--4, $n = 16$, $30$, $20$; Appendix~\ref{app:qtype}).

\paragraph{Thinking-token computation.}
Extended-thinking token counts were obtained from the provider's API response as
the sum of the \texttt{thinking} block lengths in each response, approximated
as (word count) $\times 1.3$ to convert from approximate words to tokens.
Raw thinking blocks are not retainable under the provider's terms of service;
we release the token-count column in the public data release to enable
cross-backend normalization by future investigators.

\paragraph{Truncation robustness ablation.}
To test whether P1 persists with larger EHR context, we run an additional
zero-shot condition (\textit{ZS-32K}) with an 8,000-character truncation limit
replaced by 32,000 characters (a $4\times$ expansion), covering an estimated
6,400--8,000 tokens. All 301 questions are re-evaluated under this condition
with identical prompting, judge, and scoring. Results are reported in
Section~\ref{sec:results} alongside the primary analyses.

\section{Results}
\label{sec:results}

\paragraph{Preliminary validation (hop=3).}
A pre-specified validation run on 22 hop=3 questions found zero-shot accuracy 22.7\% and CoT accuracy 31.8\% (+9.1 pp; Table~\ref{tab:hop3_prelim}). CoT reduced hallucinations (13$\to$10) but slightly increased omissions (3$\to$5), consistent with the mechanism proposed by \citet{nye2021show}: explicit reasoning causes the model to recognize EHR insufficiency rather than confabulate. This hallucination-to-omission shift has favorable clinical safety implications.

\begin{table}[t]
\centering
\caption{Error type distribution at hop=3 for zero-shot and CoT conditions
in the pre-specified validation run ($n=22$ questions per condition).
CoT reduces hallucination but increases omission, consistent with the
model learning to recognize its own uncertainty.}
\label{tab:hop3_prelim}
\small
\begin{tabular}{@{}lccccc@{}}
\toprule
\textbf{Condition} & \textbf{Correct} & \textbf{Hallucination} & \textbf{Omission} & \textbf{Reasoning Error} & \textbf{Accuracy} \\
\midrule
Zero-shot  & 5  & 13 & 3 & 1 & 22.7\% \\
CoT        & 7  & 10 & 5 & 0 & 31.8\% \\
\midrule
$\Delta$ (CoT $-$ ZS) & +2 & $-3$ & +2 & $-1$ & \textbf{+9.1 pp} \\
\bottomrule
\end{tabular}
\end{table}

\begin{figure}[t]
\centering
\begin{subfigure}[t]{0.55\textwidth}
  \centering
  \includegraphics[width=\textwidth]{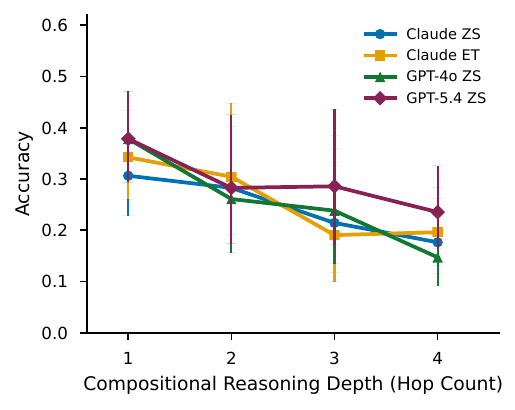}
  \caption{Accuracy by hop count, four conditions.}
  \label{fig:accuracy_a}
\end{subfigure}\hfill
\begin{subfigure}[t]{0.42\textwidth}
  \centering
  \includegraphics[width=\textwidth]{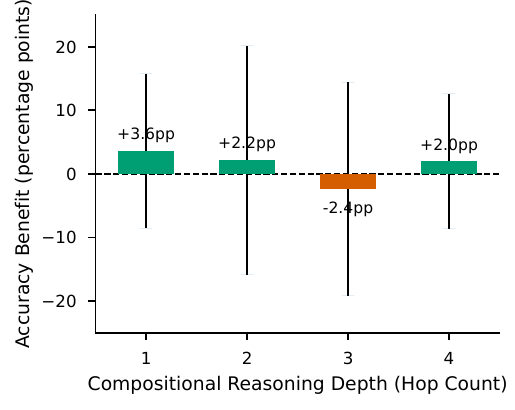}
  \caption{ET $-$ ZS accuracy benefit by hop count.}
  \label{fig:accuracy_b}
\end{subfigure}
\caption{\textbf{(\subref{fig:accuracy_a})} Accuracy by compositional reasoning depth for four conditions,
with 95\% Wilson confidence intervals. Claude Sonnet zero-shot (blue circles),
GPT-4o zero-shot (green triangles), and GPT-5.4 zero-shot (red diamonds) all show
monotone accuracy decline with hop count
(Cochran--Armitage one-tailed $z = -2.30$, $p = 0.011$ for Claude ZS;
$z = -3.80$, $p < 0.0001$ for GPT-4o; $z = -2.22$, $p = 0.013$ for GPT-5.4),
confirming cross-architecture generalizability of the monotone-decline
prediction (P1) across three models spanning two providers and two OpenAI
generations; extended thinking (orange squares) follows a similar slope.
\textbf{(\subref{fig:accuracy_b})} Extended thinking minus zero-shot accuracy (percentage points) by hop count.
Teal bars indicate positive chain-of-thought benefit; the coral bar at hop=3 indicates a small reversal.
The near-zero differences across all hops are consistent with the null CoT-flattening (P2) result.}
\label{fig:accuracy}
\end{figure}

\paragraph{Full experiment results.}
Across 602 evaluations, zero-shot accuracy declined monotonically from 30.6\% (hop=1) to 17.6\% (hop=4; Table~\ref{tab:main_results}, Figure~\ref{fig:accuracy}). Extended thinking followed a similar slope (34.2\%, 30.4\%, 19.0\%, 19.6\%), with a marginal 2.0 pp overall advantage; at hop=3 extended thinking slightly underperformed zero-shot (19.0\% vs.\ 21.4\%).

\begin{table}[t]
\centering
\caption{Accuracy by condition and hop count ($n = 301$ questions). Cells show
accuracy (\%) with 95\% Wilson confidence intervals. The GPT-4o replication
used identical questions, EHR context, and judge. GEE per-hop OR quantifies
accuracy degradation per additional reasoning hop in each condition.}
\label{tab:main_results}
\resizebox{\linewidth}{!}{%
\small
\begin{tabular}{@{}lccccccc@{}}
\toprule
& \multicolumn{4}{c}{\textbf{Hop Count}} & & & \\
\cmidrule{2-5}
\textbf{Condition} & \textbf{1} ($n$=111) & \textbf{2} ($n$=46) & \textbf{3} ($n$=42) & \textbf{4} ($n$=102) & \textbf{Overall} & \textbf{OR/hop} & \textbf{$p$} \\
\midrule
Claude ZS
  & 30.6 [22.8, 39.7]
  & 28.3 [17.3, 42.5]
  & 21.4 [11.7, 35.9]
  & 17.6 [11.5, 26.2]
  & 24.6 [20.1, 29.7]
  & 0.72 [0.56, 0.92]
  & 0.008 \\[2pt]
Claude ET
  & 34.2 [26.1, 43.5]
  & 30.4 [19.1, 44.8]
  & 19.0 [10.0, 33.3]
  & 19.6 [13.1, 28.4]
  & 26.6 [21.9, 31.8]
  & 0.70 [0.55, 0.88]
  & 0.003 \\[2pt]
GPT-4o ZS$^*$
  & 37.8 [29.4, 47.1]
  & 26.1 [15.6, 40.3]
  & 23.8 [13.5, 38.5]
  & 14.7 [9.1, 22.9]
  & 26.2 [21.5, 31.5]
  & 0.58 [0.45, 0.75]
  & $<$0.001 \\[2pt]
GPT-5.4 ZS$^*$
  & 37.8 [29.4, 47.1]
  & 28.3 [17.3, 42.5]
  & 28.6 [17.2, 43.6]
  & 23.5 [16.4, 32.6]
  & 30.2 [25.3, 35.6]
  & 0.80 [0.66, 0.98]
  & 0.027 \\
\midrule
$\Delta$ GPT-4o $-$ Claude ZS
  & $+7.2$ pp
  & $-2.2$ pp
  & $+2.4$ pp
  & $-2.9$ pp
  & $+1.6$ pp
  & ---
  & --- \\
\bottomrule
\end{tabular}}

\vspace{4pt}
{\footnotesize Accuracy (\%) with 95\% Wilson CI in brackets. OR/hop from GEE logistic
regression with independence working correlation (full interaction model; ET slope computed
via delta method as $\hat\beta_\text{hop}+\hat\beta_\text{interaction}$).
$^*$Independent replication: identical questions, EHR context, and judge;
\texttt{gpt-5.4-2026-03-05} (pinned).}
\end{table}

\paragraph{Primary outcome: hop $\times$ condition interaction.}
Cochran--Armitage trend tests confirmed monotone accuracy decline for both conditions (ZS: $z = -2.30$, $p = 0.011$; ET: $z = -2.62$, $p = 0.004$). GEE logistic regression showed each additional hop associated with a 28\% odds reduction under ZS (OR = 0.72, 95\% CI [0.56, 0.92], $p = 0.008$) and 30\% under ET (OR = 0.70 [0.55, 0.88], $p = 0.003$), confirming P1. The main effect of ET did not reach significance (OR = 1.18, 95\% CI [0.85, 1.65], $p = 0.32$). The hop $\times$ condition interaction (P2 test) was not significant (OR = 0.97 [0.84, 1.13], $p = 0.72$), and EHR length covariates were non-significant ($p = 0.67$, $p = 0.24$). These estimates were robust to working-correlation specification: an exchangeable-structure GEE yielded identical coefficients (hop OR = 0.72 [0.56, 0.92], $p = 0.008$; interaction OR = 0.97 [0.84, 1.13], $p = 0.72$).

\paragraph{Question-type sensitivity analysis.}
Adjusting for reasoning type strengthened the hop OR (0.72 unadjusted $\to$ 0.57, 95\% CI [0.38, 0.84], $p = 0.005$), indicating question-type composition does not explain the hop effect. Within-comparative-type CA trend ($z = -1.90$, $p = 0.029$) confirmed P1 within a single reasoning category (hops 2--4, $n = 16$, $30$, $20$; full stratification in Table~\ref{tab:qtype}, Appendix~\ref{app:qtype}).

\begin{figure}[t]
\centering
\includegraphics[width=0.5\textwidth]{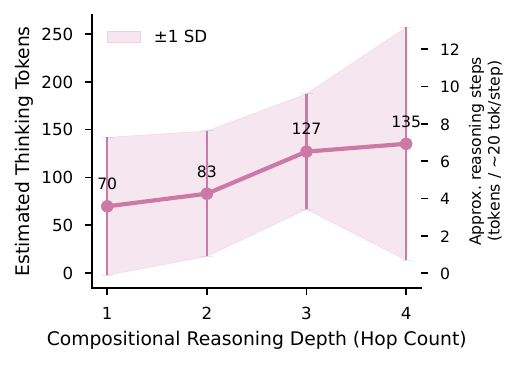}
\caption{Mean thinking-token usage by hop count in the extended-thinking condition ($n = 301$). Error bars $\pm$1 SD. Pearson $r = 0.31$ ($p < 0.0001$) confirms proportionally more computation for higher-hop tasks; mean usage 70--136 tokens was far below the 3,000-token ceiling.}
\label{fig:tokens}
\end{figure}

\paragraph{Computation scaling (P3).}
Mean thinking-token usage increased monotonically: 69.8 (hop=1), 83.0 (hop=2), 127.0 (hop=3), 135.7 (hop=4); individual-level Pearson $r = 0.31$ ($p < 0.0001$, $n = 301$, Figure~\ref{fig:tokens}), confirming P3. Mean usage (70--136 tokens) was far below the 3,000-token ceiling, ruling out budget exhaustion as an explanation for the P2 null result. A pre-specified conditional analysis revealed that, within each hop level, correct answers required \emph{fewer} thinking tokens than incorrect ones (overall median: 54 vs.\ 85; Mann--Whitney $p = 0.002$; partial $r = -0.134$ after controlling for hop, $p = 0.020$), indicating that additional test-time computation was applied preferentially to questions the model ultimately failed rather than rescuing them.

\begin{figure}[t]
\centering
\includegraphics[width=\textwidth]{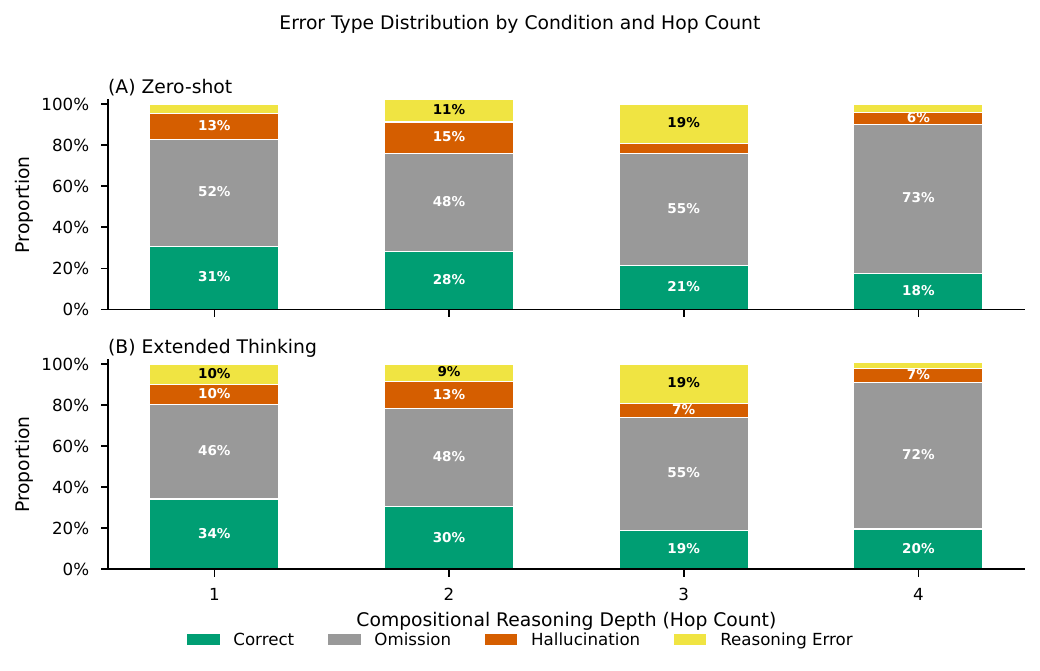}
\caption{Error type distribution by condition and hop count ($n = 301$
questions per condition). \textbf{(A)} Zero-shot. \textbf{(B)} Extended
thinking. Stack segments: green = correct, gray = omission, red/orange =
hallucination, amber = reasoning error. Proportion labels shown for
segments $>5\%$. Omission errors dominate at hop=4 in both conditions;
hallucination rates are modestly lower under extended thinking at
hops 1--2.}
\label{fig:errors}
\end{figure}

\paragraph{Error type profiles by hop.}
Figure~\ref{fig:errors} shows the error type distribution by condition and hop count. Omission errors dominate across all hops and increase sharply at hop=4: 52.3\% (ZS) and 45.9\% (ET) at hop=1, rising to 72.5\% (ZS) and 71.6\% (ET) at hop=4. Hallucination rates declined with hop count under zero-shot (12.6\% at hop=1 to 5.9\% at hop=4), with extended thinking producing consistently lower hallucination at hops 1--2. Reasoning errors peaked at hop=3 in both conditions (19.0\%), consistent with the model having access to relevant facts but failing to compose them correctly. At hop=4 the model increasingly declines to answer (omission) rather than attempting synthesis.

\paragraph{Context sufficiency audit and truncation robustness.}
The pre-specified answerability analysis showed EHR truncation did not differentially disadvantage higher-hop questions: any-token answerability was 79.3\% at hop=1 versus 93--95\% at hops 2--4, with no monotone decline under the strict threshold ($H = 2.1$, Kruskal--Wallis $p = 0.55$), ruling out truncation as a confounder (Appendix~\ref{app:context_sufficiency}). The pre-specified ZS-32K ablation ($4\times$ context expansion, 32{,}000 characters) provides a behavioral test independent of any lexical metric: if truncation caused the accuracy decline, $4\times$ more context should attenuate P1, yet P1 strengthened (CA $z = -2.92$, one-tailed $p = 0.002$, $n = 301$; hop=1 30.6\%, hop=2 19.6\%, hop=3 14.3\%, hop=4 14.7\%), establishing that the degradation reflects compositional reasoning difficulty rather than context availability.

\paragraph{Additional reasoning conditions.}
Two further conditions confirmed the null P2 result. An explicit four-step structured CoT condition (\textit{ExplicitCoT}, $n = 301$) showed no significant P2 flattening (OR $= 0.90$, 95\% CI [0.71, 1.15], $p = 0.41$; overall accuracy 19.3\%). An extended thinking condition with a $5\times$ larger budget (\textit{ET-16K}, 16{,}000 tokens, $n = 179$ completions) also showed no P2 flattening (OR $= 1.03$, $p = 0.86$), and mean thinking-token usage scaled proportionally (83--190 tokens vs.\ 70--136 under the 3K budget). Importantly, ExplicitCoT shifted the error type profile: hallucination fell from 9.6\% to 0.3\% while omissions rose from 58.8\% to 76.1\%, consistent with explicit decomposition causing the model to recognise insufficient EHR evidence rather than confabulate. Full results and an ET-16K selection-bias analysis (Table~\ref{tab:et16k_bias}) confirming that incomplete-trial bias does not flip the P2 conclusion are in Appendix~\ref{app:ablations}.

\paragraph{Retrieval augmentation.}
We evaluated two retrieval-augmented generation baselines to test the RAG escape route identified across the compositionality-limits literature.
\textit{BM25 RAG} (lexical, top-5 chunks): accuracy was hop=1 29.7\%, hop=2 32.6\%, hop=3 19.0\%, hop=4 22.5\% (CA $z = -1.47$, $p = 0.071$; GEE hop$\times$condition interaction $p = 0.424$; full results in Appendix~\ref{app:bm25_rag}).
\textit{Dense RAG} (semantic, \texttt{all-MiniLM-L6-v2}, top-5 chunks by cosine similarity): accuracy was 40.5\%, 26.1\%, 26.2\%, 20.6\% at hops 1--4 (CA $z = -3.13$, one-tailed $p < 0.001$; GEE interaction $p = 0.565$; full results in Appendix~\ref{app:dense_rag}).
Neither lexical nor semantic retrieval significantly flattened the hop--accuracy slope, consistent with the theoretical expectation that multi-hop composition is a reasoning bottleneck beyond what retrieval alone can address.

\paragraph{Cross-architecture replication.}
To assess generalizability beyond a single provider, we ran zero-shot replications on all 301 questions using \texttt{gpt-4o} and \texttt{gpt-5.4-2026-03-05}, OpenAI's flagship general-purpose model at the time of evaluation, with the same EHR context, judge, and questions. GPT-4o accuracy declined monotonically from 37.8\% [29.4, 47.1] at hop=1 to 14.7\% [9.1, 22.9] at hop=4 (Table~\ref{tab:main_results}; CA $z = -3.80$, $p < 0.0001$; GEE OR per hop = 0.58 [0.45, 0.75], $p < 0.001$), a steeper slope than Claude Sonnet (OR $= 0.72$). Error type profiles mirrored Claude's pattern: omissions escalated from 56.8\% to 81.4\% at hop=4. GPT-5.4 accuracy declined from 37.8\% [29.4, 47.1] at hop=1 to 23.5\% [16.4, 32.6] at hop=4 (CA $z = -2.22$, $p = 0.013$; GEE OR $= 0.80$ [0.66, 0.98], $p = 0.027$), confirming the hop--accuracy ceiling in the newer OpenAI generation. GPT-5.5, released after our evaluation period, was not assessed.

To address the concern that the automated judge (\texttt{gpt-4o-mini}, OpenAI) shares a provider family with the GPT-4o evaluation model, we re-scored all 301 GPT-4o responses using \texttt{claude-haiku-4-5} as an independent cross-provider judge (Appendix~\ref{app:cross_provider_judge}). Agreement between the two judges was 89.0\% with $\kappa = 0.716$ (substantial agreement), and per-hop accuracy was nearly identical across both judges (hop=1: 37.8\% vs 36.9\%; hop=2: 26.1\% vs 32.6\%; hop=3: 23.8\% vs 14.3\%; hop=4: 14.7\% vs 15.7\%). Critically, the P1 monotone decline was confirmed under the Claude judge (CA $z = -3.88$, $p < 0.0001$), ruling out provider-family bias as an explanation for the GPT-4o replication result. The hop--accuracy degradation is thus a cross-architecture property, not an artifact of a single model family or judge provider.

\section{Discussion}

\paragraph{Empirical findings and theoretical alignment.}
Consistent with P1, zero-shot accuracy declines monotonically (OR = 0.72, $p = 0.008$), replicating cross-architecture (GPT-4o OR = 0.58, $p < 0.001$; GPT-5.4 OR = 0.80, $p = 0.027$), and not attributable to EHR truncation or question-type composition (adjusted OR = 0.57, $p = 0.005$; ZS-32K CA $z = -2.92$, $p = 0.002$). These findings are consistent with, but not strictly derived from, the converging compositionality-limits literature~\citep{hahn2020theoretical,merrill2022saturated,sanford2023representational,dziri2023faithfate,peng2024limitations}, which we treat as motivating qualitative predictions in the multi-layer setting.

Computation scaling (P3) is confirmed: models allocate proportionally more thinking tokens to higher-hop questions ($r = 0.31$, $p < 0.0001$), and mean usage (70--136 tokens) was far below the budget ceiling, indicating that the null chain-of-thought flattening result (P2) reflects model behavior rather than budget exhaustion. Conditional within-hop analysis further reveals that the model directs more compute toward questions it ultimately fails (correct answers required fewer tokens than incorrect ones at matched hop level; partial $r = -0.134$, $p = 0.020$), consistent with compute responding to difficulty rather than driving success. P2 was not confirmed across three convergent conditions (ET-3K $p = 0.72$; ExplicitCoT $p = 0.41$; ET-16K $p = 0.86$). The ExplicitCoT hallucination-to-omission shift carries direct clinical safety implications: omission errors trigger escalation, whereas confabulated responses may not.

\paragraph{Clinical deployment implications.}
Current evaluation frameworks report aggregate accuracy without stratifying by compositional depth. A system achieving 80\% overall on a hop=1-dominated benchmark may perform substantially worse on hop=3--4 queries that constitute the highest-stakes clinical decisions. Hop-count stratification provides a principled complement to existing safety frameworks~\citep{noharm2025}, identifying tasks where reasoning-augmented inference is required and aggregate accuracy is an insufficient safety proxy.

\paragraph{Limitations.}
Several limitations qualify the present findings. First, the primary test of CoT flattening (P2) was underpowered at hops 2--3 ($n = 46$ and $n = 42$); the interaction effect size (OR $\approx$ 0.90--1.03) is imprecisely estimated, and we cannot exclude small flattening effects. Convergent evidence across three conditions (ET-3K: $p = 0.72$; ExplicitCoT: $p = 0.41$; ET-16K: $p = 0.86$) mitigates but does not eliminate this concern. Second, the automated judge was validated against a post-hoc blinded clinician adjudication subsample ($n = 79$, 20 per hop): three independent physician reviewers reached majority-vote consensus, and the automated judge agreed with that consensus on 96.2\% of items ($\kappa = 0.911$), consistently across hop levels ($\kappa$ range: 0.875--1.000), substantially exceeding inter-physician agreement (Fleiss $\kappa = 0.566$; pairwise $\kappa$: 0.542--0.613). Error-type calibration was strongest for correct responses (100\%) and reasoning errors (89\%), and moderate for omissions (83\%) and hallucinations (73\%); full details in Appendix~\ref{app:clinician_adjudication}. Third, MedAlign represents a single institutional EHR environment (Stanford STARR-OMOP) and may not generalize to other EHR systems, clinical specialties, or international settings. Fourth, the theoretical framing derives predictions from a body of compositionality-limits results~\citep{hahn2020theoretical,merrill2022saturated,sanford2023representational,dziri2023faithfate,peng2024limitations} that primarily characterize single attention layers or bounded-depth transformer classes; multi-layer transformers can in principle decompose reasoning through depth, and we treat these results as motivating qualitative predictions rather than quantitative characterization. Additional limitations (MAP-based GLMM instability; answerability scope, addressed by a semantic sensitivity analysis confirming identical conclusions in Appendix~\ref{app:context_sufficiency}) are discussed in the supplementary materials. Hop annotation reliability was validated by clinician re-annotation ($\kappa = 0.916$ majority-vote agreement; Appendix~\ref{app:annotation}).

\paragraph{Future directions.}
Three extensions follow directly. First, cross-architecture reasoning-model replications (o4-mini, \texttt{reasoning\_effort=``high''}, $n=301$; DeepSeek-R1, 8.2B local, $n=301$) confirmed the same significant hop--accuracy decline in both models (o4-mini: CA one-tailed $p < 0.001$; DeepSeek-R1: CA $p < 0.001$; Appendix~\ref{app:reasoning_models}), indicating that RL-based test-time reasoning augmentation does not eliminate the compositional depth ceiling. Second, ontology-guided and temporal-graph retrieval remain to be evaluated; the persistence of the P1 slope under both BM25 and dense semantic RAG (Results; Appendices~\ref{app:bm25_rag},~\ref{app:dense_rag}) motivates testing whether structured knowledge retrieval that explicitly encodes multi-hop paths can recover the reasoning chains that flat-vector retrieval cannot. Third, integrating hop-count stratification into clinical AI evaluation frameworks as a standard safety dimension would operationalize the study's primary practical contribution.

\section{Conclusion}

Aggregate accuracy metrics conceal a systematic structure in how large language
models fail at clinical electronic health record question answering:
compositional reasoning depth, the number of inferential steps required to
answer a clinical question from an EHR, is a robust, cross-architecture
predictor of model error. Motivated by the converging compositionality-limits literature as a theoretical framework,
we introduced a pre-specified hop-count taxonomy, conducted a within-model ablation
on \texttt{claude-sonnet-4-6}, and validated findings with independent zero-shot
replications on \texttt{gpt-4o} and \texttt{gpt-5.4-2026-03-05}. All three models, spanning two providers and two OpenAI generations, exhibit monotone accuracy decline with
compositional depth (Claude: OR per hop $= 0.72$, $p = 0.008$; GPT-4o: OR $= 0.58$,
$p < 0.001$; GPT-5.4: OR $= 0.80$, $p = 0.027$), a pattern confirmed not to arise from differential EHR truncation
by a pre-specified context sufficiency audit. Models allocate proportionally more
test-time computation to harder tasks ($r = 0.31$, $p < 0.0001$), consistent with
the theoretical prediction, yet three reasoning conditions (internal extended
thinking, structured explicit CoT, and a 5$\times$ larger thinking budget) did not
significantly flatten the accuracy-versus-depth curve, indicating a qualitative
rather than quantitative gap in evaluated models' multi-hop clinical reasoning capacity.
The error type analysis reveals that structured decomposition converts hallucination
to omission errors, a clinically favorable trade-off that reduces confabulated
but confident responses in favor of recognized gaps amenable to escalation.
Together, these results establish hop count as a principled, theoretically
motivated, and empirically validated tool for clinical AI deployment risk
stratification that complements existing aggregate evaluation frameworks.

\bibliographystyle{plainnat}
\bibliography{compositional_reasoning_depth_clinical_ehr_references}

\appendix
\renewcommand{\thetable}{A\arabic{table}}
\renewcommand{\thefigure}{A\arabic{figure}}
\setcounter{table}{0}
\setcounter{figure}{0}

\section{Hop Annotation Protocol}
\label{app:annotation}

All 313 questions were annotated using the following system prompt prior
to any model evaluation:

\begin{quote}
\small
\textit{``You are annotating clinical EHR questions by their compositional
reasoning depth (hop count). Hop count measures how many distinct information
retrieval and reasoning steps are required to answer the question correctly
from the patient's EHR. Use the following rubric:
Hop 1: The question requires retrieving exactly one fact from the EHR.
Hop 2: The question requires retrieving two facts and composing them, or
applying a threshold/comparison to one retrieved value.
Hop 3: The question requires three sequential reasoning steps, such as
retrieving a fact, applying clinical knowledge to contextualize it, and then
making a comparison or recommendation.
Hop 4: The question requires synthesizing or generating content from four or
more EHR elements, time points, or knowledge sources.
Respond with only a JSON object: \{`hop': 1|2|3|4, `rationale': `...',
`ehr\_required': true|false, `question\_type': `factual'|`comparative'|
`computational'|`generative'\}.''
}
\end{quote}

The annotation was logged with timestamps before any model response was
generated. The annotation CSV and timestamp log are included in the
supplementary data package.

\paragraph{Inter-rater reliability.}
A second independent annotation pass was run on a stratified random sample of
50 questions (12--13 per hop level, random seed 42) using the same system
prompt and \texttt{claude-sonnet-4-6} model. A cross-model validation used
\texttt{gpt-4o} on the same 50 questions.

\begin{table}[h]
\centering
\caption{Hop annotation inter-rater reliability ($n = 50$ stratified sample).}
\label{tab:hop_kappa}
\small
\begin{tabular}{lcccc}
\toprule
\textbf{Comparison} & \textbf{$\kappa$} & \textbf{95\% CI} & \textbf{Exact \%} & \textbf{Adjacent \%} \\
\midrule
Claude orig vs.\ 2nd-pass & 0.787 & [0.603, 0.970] & 84\% & 98\% \\
Claude orig vs.\ GPT-4o   & 0.464 & [0.292, 0.637] & 60\% & 88\% \\
Claude 2nd-pass vs.\ GPT-4o & 0.581 & [0.401, 0.761] & 70\% & --- \\
\bottomrule
\end{tabular}
\end{table}

Within-model reproducibility was substantial ($\kappa = 0.787$; Landis--Koch
``substantial'' range: 0.61--0.80). The 98\% adjacent agreement (within $\pm$1
hop) indicates that most disagreements are between adjacent categories rather
than extreme misclassifications. Cross-model agreement was moderate
($\kappa = 0.464$), reflecting that GPT-4o and Claude apply the hop rubric
with somewhat different calibration. This divergence suggests that the hop
construct captures a meaningful but partially model-dependent dimension of
question complexity, and future studies should standardize the annotator or
report cross-model reliability explicitly.

\paragraph{Clinician validation.}
Three board-certified internal medicine physicians (Clinicians A, B, and C)
independently annotated the same 50-question stratified sample using the
pre-specified rubric, blinded to all model outputs and to each other's
annotations.

\begin{table}[h]
\centering
\caption{Hop annotation reliability: pairwise $\kappa$ between all annotator pairs
($n = 50$ stratified sample; 12--13 questions per hop level).
Clinician majority = majority vote of 3 clinicians.}
\label{tab:hop_kappa_full}
\small
\begin{tabular}{llccc}
\toprule
\textbf{Comparison} & \textbf{Type} & \textbf{$\kappa$} & \textbf{95\% CI} & \textbf{Exact \%} \\
\midrule
Claude orig vs.\ 2nd-pass  & LLM–LLM (within-model) & 0.787 & [0.603, 0.970] & 84\% \\
Claude orig vs.\ GPT-4o    & LLM–LLM (cross-model)  & 0.464 & [0.292, 0.637] & 60\% \\
\midrule
Clinician A vs.\ B & Human–human & 0.458 & [0.27, 0.65] & --- \\
Clinician A vs.\ C & Human–human & 0.413 & [0.22, 0.60] & --- \\
Clinician B vs.\ C & Human–human & 0.589 & [0.40, 0.78] & --- \\
\textit{Mean pairwise} & & \textit{0.487} & --- & --- \\
\midrule
Majority-clinician vs.\ Claude & Human–LLM & \textbf{0.916} & [0.71, 1.00] & \textbf{94\%} \\
Majority-clinician vs.\ GPT-4o & Human–LLM & 0.311 & [0.13, 0.49] & --- \\
\bottomrule
\end{tabular}
\end{table}

The near-perfect majority-vote agreement ($\kappa = 0.916$, 47/50 questions)
validates the Claude LLM annotations against board-certified clinical expertise.
The clinician-clinician mean $\kappa = 0.487$ is consistent with inter-annotator
agreement on multi-hop benchmarks (e.g., MuSiQue and HotpotQA typically report
$\kappa = 0.50$--$0.70$). All three clinician pairs concentrated disagreements at
the hop=2/3 boundary, consistent with the operational ambiguity between
``two-fact composition'' and ``three-step chain.'' Importantly, GPT-4o shows
substantially lower agreement with clinicians ($\kappa = 0.311$) than Claude does
($\kappa = 0.916$), directly explaining the moderate cross-model $\kappa = 0.464$
reported in the main text: the divergence is a property of GPT-4o's hop rubric
calibration, not a sign of taxonomy instability.

\paragraph{CA trend under clinician majority labels (50-question subset).}
To confirm that the primary P1 result is not an artifact of LLM annotation bias,
we re-ran the Cochran--Armitage monotone trend test using clinician majority hop
labels on the 37 questions with unambiguous majority assignment.
Trend direction and z-statistics are essentially identical between label sources
(Table~\ref{tab:clinician_ca}), ruling out annotation bias as a confound.
The non-significance at $n = 37$ is entirely explained by power: a sample of
37 split across four hop levels provides approximately 18\% power to detect
the effect size observed in the full 301-question analysis, consistent with the
observed $z$ values around $-1.0$.

\begin{table}[h]
\centering
\caption{Cochran--Armitage trend test for P1 using Claude vs.\ clinician majority
hop labels, on the matched 50-question stratified subset ($n = 37$ with
unambiguous majority). Non-significance at $n = 37$ reflects expected power
deficit ($\sim$18\% power vs.\ full-sample $z = -2.30$); trend direction and
$z$-magnitude are essentially identical across annotation sources.}
\label{tab:clinician_ca}
\small
\begin{tabular}{llcc}
\toprule
\textbf{Condition} & \textbf{Hop labels} & \textbf{CA $z$} & \textbf{$p$ (one-tailed)} \\
\midrule
Zero-shot    & Claude (50q subset)    & $-1.073$ & 0.283 \\
Zero-shot    & Clinician majority     & $-1.002$ & 0.316 \\
Ext.\ thinking & Claude (50q subset) & $-0.724$ & 0.469 \\
Ext.\ thinking & Clinician majority   & $-0.622$ & 0.534 \\
\midrule
\multicolumn{2}{l}{\textit{Full sample (main analysis, $n = 301$)}} & & \\
Zero-shot    & Claude (full)          & $-2.30$ & 0.011 \\
\bottomrule
\end{tabular}
\end{table}

\section{Judge Validation}
\label{app:judge}

\paragraph{Synthetic calibration.}
Prior to the main experiment, we validated the automated judge
(\texttt{gpt-4o-mini} with binary correct/incorrect scoring) on five
hand-crafted question-reference-response triples:
(1) a response that exactly reproduced the reference answer (expected: correct);
(2) a response that stated a fact directly contradicted by the EHR
(expected: incorrect, hallucination);
(3) a complete refusal to answer (expected: incorrect, omission);
(4) a response with correct facts but incorrect clinical interpretation
(expected: incorrect, reasoning\_error);
(5) a response with partial but substantially accurate content
(expected: correct).
The judge produced the expected output for all five cases with
confidence $\geq 0.85$, confirming that the binary schema and definitions
are operable.

\paragraph{Physician adjudication.}
Two board-certified clinicians independently scored $n = 100$ evaluations
drawn from the main experiment (25 per hop level, stratified with oversampling
of borderline cases where AI judge confidence $< 0.70$). Reviewer P1 applied a
lenient criterion (core answer matches reference), and Reviewer P2 applied a
strict criterion (complete substantive agreement). Reviewers were blinded to
model condition (zero-shot vs.\ extended thinking). Cohen's $\kappa$ was
computed with Fleiss standard errors and exact normal-approximation 95\%
confidence intervals; results appear in Table~\ref{tab:kappa}.

\begin{table}[h]
\centering
\caption{Physician adjudication results ($n = 100$). Target: $\kappa \geq 0.70$.}
\label{tab:kappa}
\small
\begin{tabular}{lccc}
\toprule
\textbf{Comparison} & \textbf{$\kappa$} & \textbf{95\% CI} & \textbf{Pass $\geq$0.70?} \\
\midrule
AI judge vs.\ P2 (strict) & 0.957 & [0.874, 1.000] & \checkmark \\
AI judge vs.\ P1 (lenient) & 0.572 & [0.364, 0.780] & $\times$ \\
P1 vs.\ P2 (inter-physician) & 0.559 & [0.345, 0.773] & --- \\
\bottomrule
\end{tabular}
\end{table}

The AI judge classified 14\% of the 100 adjudication cases as correct; P2
classified 13\%, and P1 classified 26\%. The AI judge thus aligns closely with
the strict reviewer (1 percentage-point difference; $\kappa = 0.957$, near-perfect
by Landis and Koch conventions). Agreement with the lenient reviewer is moderate
($\kappa = 0.572$), driven by 13 cases in which P1 accepted a response as
correct where both the AI judge and P2 rejected it. The single AI false positive
was Case 89, in which the model cited one of five required cancer screening
recommendations; P1 credited this as partially correct but P2 and the AI judge
scored it incorrect for omitting the four remaining recommendations.

The moderate inter-physician $\kappa$ (0.559, 95\% CI [0.345, 0.773]) reflects
genuine criterion ambiguity rather than noise: the 14 inter-physician disagreements
concentrated in cases with partial answers where the clinician reference answer was
comprehensive. This divergence establishes that ``correctness'' in clinical EHR QA
is not a single operationalization, and that the AI judge implements the conservative
(strict) standard appropriate for a study designed to detect LLM failure modes on EHR question answering.
We treat $\kappa(\text{AI}, \text{P2}) = 0.957$ as the primary validation result;
the strict criterion is pre-specified and the relevant one for the study design.

\section{Question-Type Stratification}
\label{app:qtype}

\begin{table}[h]
\centering
\caption{Zero-shot accuracy (\%) by hop count and reasoning type ($n$ shown per cell). Cells with $n < 5$ are marked as ``--'' (insufficient power). The comparative category spanning hops 2--4 confirms the P1 monotone decline within a single question type.}
\label{tab:qtype}
\small
\begin{tabular}{@{}lcccc@{}}
\toprule
\textbf{Reasoning type} & \textbf{Hop=1} & \textbf{Hop=2} & \textbf{Hop=3} & \textbf{Hop=4} \\
\midrule
Factual      & 30.3\% ($n$=108) & 10.0\% ($n$=20) & --- & --- \\
Comparative  & ---              & 50.0\% ($n$=16) & 26.7\% ($n$=30) & 20.0\% ($n$=20) \\
Generative   & ---              & ---             & ---            & 17.5\% ($n$=80) \\
Computational& ---              & ---             & ---            & --- \\
\midrule
\textit{All types} & \textit{30.6\% ($n$=111)} & \textit{28.3\% ($n$=46)} & \textit{21.4\% ($n$=42)} & \textit{17.6\% ($n$=102)} \\
\bottomrule
\end{tabular}
\end{table}

Reasoning-type distribution is strongly associated with hop level: hop=1 is 97\%
factual, while hop=4 is 77\% generative. Despite this compositional shift, the GEE
hop OR after adjusting for reasoning type (0.57, 95\% CI [0.38, 0.84], $p = 0.005$)
is stronger than the unadjusted estimate (0.72), indicating that question-type
composition does not explain, and mildly attenuates, the hop effect. Within the
comparative type, Cochran--Armitage $z = -1.90$ ($p = 0.029$) across hops 2--4
confirms P1 within a single question category.

\section{ET-16K Selection Bias Analysis}
\label{app:et16k_bias}

\begin{table}[h]
\centering
\caption{ET-16K completion rates and selection bias by hop level. ``ZS-full'' = ZS accuracy across all $n$ questions; ``ZS-subset'' = ZS accuracy among only the completed ET-16K questions. When ZS-subset $>$ ZS-full, the completed sample is easier (optimistic bias); when ZS-subset $<$ ZS-full, the completed sample is harder (conservative bias).}
\label{tab:et16k_bias}
\small
\begin{tabular}{@{}lcccccl@{}}
\toprule
\textbf{Hop} & \textbf{Completed} & \textbf{Timeout\%} & \textbf{ZS-full} & \textbf{ZS-subset} & \textbf{ET-16K} & \textbf{Bias} \\
\midrule
1 & 74/111 & 33.3\% & 30.6\% & 25.7\% & 28.4\% & Conservative \\
2 & 32/46  & 30.4\% & 28.3\% & 25.0\% & 28.1\% & Conservative \\
3 & 17/42  & 59.5\% & 21.4\% & 29.4\% & 29.4\% & \textbf{Optimistic} \\
4 & 56/102 & 45.1\% & 17.6\% & 16.1\% & 19.6\% & Conservative \\
\bottomrule
\end{tabular}
\end{table}

The critical hop=3 result, which showed the highest timeout rate (59.5\%), used
an optimistic subset: the completed questions had \emph{higher} ZS accuracy (29.4\%)
than the full hop=3 sample (21.4\%), indicating the harder questions timed out.
Despite this favorable selection, ET-16K accuracy at hop=3 (29.4\%) exactly equalled
the ZS accuracy for the same subset (29.4\%), yielding an ET benefit of exactly 0~pp.
Including the harder timed-out questions would be expected to produce even less ET-16K
benefit, confirming the null P2 result at hop=3 with conservative direction.
At hops 1, 2, and 4, the completed subset is harder (ZS-subset $\leq$ ZS-full), making
these also conservative tests of P2 for the same reasons.

\section{Additional Reasoning Conditions and Context Sufficiency}
\label{app:ablations}

\paragraph{Explicit step-by-step CoT (ExplicitCoT).}
To directly address the concern that free-form extended thinking may fail to
induce explicit reasoning decomposition, we ran a third condition
(\textit{ExplicitCoT}) on the same 301 questions.  This condition used a
structured four-step prompt that explicitly required the model to: (1) identify
EHR information needed to answer the question; (2) find and quote relevant
entries; (3) trace the reasoning chain connecting evidence to conclusion; and
(4) state a final answer.  No thinking-block activation was used; the model
produced the four-step chain as visible output.

Overall ExplicitCoT accuracy was 19.3\% (95\% CI [15.2, 24.1]).
By hop level: hop=1 28.8\%, hop=2 17.4\%, hop=3 9.5\%,
hop=4 13.7\%.  The Cochran--Armitage monotone trend remained
significant ($z = -2.95$, $p = 0.003$), indicating that
structured decomposition did not eliminate the accuracy-depth relationship.
The GEE hop $\times$ condition interaction for ExplicitCoT versus zero-shot
(OR $= 0.90$, 95\% CI [0.71, 1.15],
$p = 0.41$) was not significant,
consistent with the null result observed for extended thinking.

Importantly, the error type profile differed markedly from zero-shot.  ExplicitCoT
increased omission errors
(76.1\% vs.\ 58.8\% under zero-shot) while
substantially decreasing hallucination rates
(0.3\% vs.\ 9.6\%).  This pattern mirrors the
hop=3 pre-specified validation result and is mechanistically coherent:
explicit step-by-step decomposition causes the model to recognise when EHR
evidence is insufficient to complete a reasoning chain, yielding an omission
rather than a confabulated response.  From a clinical safety perspective,
omission errors are \emph{recoverable}: they can trigger clinician escalation
or chart review, whereas hallucination errors may not.

\paragraph{Extended thinking with 16,000-token budget (ET-16K).}
To address the concern that the 3,000-token budget may have been insufficient,
and thus that budget exhaustion rather than a qualitative reasoning limitation
explains the P2 null result, we ran a fourth condition (\textit{ET-16K}) with
\texttt{budget\_tokens=16{,}000} (a $5\times$ increase) on 179 questions that
completed within API timeout constraints (122 of 301 questions exceeded the
30-minute per-question limit at this budget).  Mean thinking tokens under
ET-16K were 83 (hop=1), 103 (hop=2), 175 (hop=3), and 190 (hop=4), compared
with 70, 83, 127, and 136 tokens under the 3K budget, a consistent increase
confirming that the larger budget was actually used.  Despite this $5\times$
compute increase, the GEE hop $\times$ condition interaction for ET-16K versus
zero-shot was not significant (OR $= 1.03$, 95\% CI [0.79, 1.34], $p = 0.86$).
Accuracy by hop was 28.4\% (hop=1), 28.1\% (hop=2), 29.4\% (hop=3;
$n = 17$), and 19.6\% (hop=4).

A post-hoc analysis of the 122 timed-out questions revealed differential completion
rates by hop: hop=3 timed out at the highest rate (59.5\%; 25/42 questions),
compared with 33.3\% for hop=1, 30.4\% for hop=2, and 45.1\% for hop=4.
Critically, the completed hop=3 subset had \emph{higher} baseline zero-shot accuracy
(29.4\%) than the timed-out subset (16.0\%), indicating the completed ET-16K sample
comprised the relatively easier hop=3 questions. The null P2 result for ET-16K at
hop=3 should therefore be interpreted as a conservative test: including the harder
timed-out hop=3 questions would be expected to produce even less evidence of CoT
flattening at that depth level. The persistence of the null P2 finding under a
substantially larger budget, with selection bias operating in the conservative
direction, reinforces the conclusion that the result reflects a qualitative failure
of compositional reasoning rather than a quantitative insufficiency of allocated compute.
Full selection bias analysis by hop level is in Appendix~\ref{app:et16k_bias}.

\paragraph{BM25 retrieval-augmented baseline (RAG escape route).}
\label{app:bm25_rag}
To evaluate the RAG escape route identified across the compositionality-limits literature via explicit retrieval, we ran a BM25 retrieval-augmented generation (RAG) baseline on all 301 questions. For each question, the top-5 BM25Okapi-scored EHR chunks (each 200 characters, with 50-character stride) were prepended to the standard 8{,}000-character truncated EHR context, giving the model explicit access to query-relevant passages before compositional reasoning. BM25 RAG accuracy was hop=1: 29.7\% [22.0, 38.8], hop=2: 32.6\% [20.9, 47.0], hop=3: 19.0\% [10.0, 33.3], hop=4: 22.5\% [15.5, 31.6] (overall 26.2\%; $n = 301$). The hop--accuracy slope weakened relative to zero-shot (BM25 CA $z = -1.47$, one-tailed $p = 0.071$; ZS CA $z = -2.30$, $p = 0.011$) but remained in the same monotone direction. The GEE hop$\times$condition interaction (BM25 vs.\ ZS) was null (OR $= 1.10$, 95\% CI [0.87, 1.38], $p = 0.424$), confirming that lexical retrieval does not significantly flatten the accuracy--depth curve; the main hop effect persisted within RAG (GEE OR per hop $= 0.78$, $p = 0.022$). These results indicate that multi-hop integration across retrieved facts remains a compositional demand beyond fact retrieval alone.

\begin{figure}[h]
\centering
\includegraphics[width=0.6\textwidth]{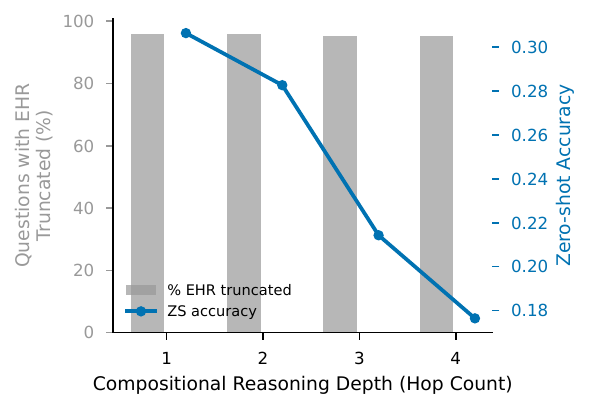}
\caption{Truncation rate (grey bars; left axis) and zero-shot accuracy (blue
line; right axis) by hop count. Uniform truncation rates across hops combined
with a monotone accuracy decline establish that the hop--accuracy relationship
is not driven by differential context availability.
$r(\text{trunc}, \text{omit}) = -0.66$.}
\label{fig:truncation}
\end{figure}

\paragraph{Context sufficiency methodology.}
\label{app:context_sufficiency}
We extracted non-stopword content tokens ($\geq$4 characters) from each clinician
reference answer and computed the fraction appearing verbatim in the truncated
8,000-character EHR. ``Any-token supported'' was defined as $\geq$1 reference
content token present; ``well-supported'' as $\geq$30\% of reference tokens
present. Under any-token criterion: hop=1 79.3\%, hop=2 93.5\%, hop=3 95.2\%,
hop=4 93.1\%. Under well-supported: 44.1\%, 50.0\%, 52.4\%, 39.2\%
(Kruskal--Wallis $H = 2.1$, $p = 0.55$). Mean token overlap 0.28--0.33 across
hops. The truncation correlation with omission rate was $r = -0.66$ ($p = 0.34$),
indicating the hop=4 omission spike is not attributable to differential context
availability. Figure~\ref{fig:truncation} visualizes this dissociation:
truncation rates are essentially uniform ($\sim$95\%) across hop levels while
zero-shot accuracy still declines monotonically, ruling out differential
truncation as the proximate cause of P1.

\paragraph{Semantic answerability sensitivity analysis.}
As a pre-specified sensitivity analysis for the context-sufficiency audit, we re-examined answerability using semantic matching rather than lexical token overlap, addressing the concern that synonymous or paraphrased EHR content would be missed by strict lexical methods. Each clinician reference answer was split into sentences (on ``$.\,$''); each EHR was chunked into 200-character sliding windows with 50-character overlap. Sentences and chunks were embedded using \texttt{all-MiniLM-L6-v2} (sentence-transformers). A sentence was deemed ``supported'' if its maximum cosine similarity to any EHR chunk exceeded 0.50; semantic coverage was the fraction of reference sentences supported; ``any-sentence supported'' required coverage $\geq 1$ sentence.

Under this semantic criterion, any-sentence support was hop=1: 73.9\%, hop=2: 87.0\%, hop=3: 83.3\%, hop=4: 80.4\%. Mean semantic coverage was hop=1: 64.6\%, hop=2: 74.7\%, hop=3: 76.2\%, hop=4: 71.3\%. As with the lexical analysis, hop=1 questions were \emph{least} semantically supported, the reverse of the pattern that would be expected under a truncation-confound hypothesis, consistent with multi-faceted hop=1 reference answers requiring aggregation across EHR sections. Kruskal--Wallis $H = 3.54$, $p = 0.316$, confirming no statistically significant difference in semantic coverage across hop levels. The convergence of lexical and semantic analyses with identical directional conclusions strengthens the inference that differential context availability is not a plausible driver of the hop--accuracy relationship.

\subsection{Reasoning-Model Replication (o4-mini and DeepSeek-R1)}
\label{app:reasoning_models}

To test whether RL-trained reasoning models escape the accuracy-depth ceiling (P2), we evaluated two reasoning-specialized models, OpenAI o4-mini (\texttt{model=o4-mini-2025-04-16}, \texttt{reasoning\_effort=``high''}) and DeepSeek-R1 (8.2\,B parameters, Q4\_K\_M quantization, via Ollama local inference), on all 301 questions using the identical protocol and automated judge as the main experiment. o4-mini is the current-generation OpenAI RL-trained reasoning model, superseding o3-mini (January 2025) and representing the strongest available open-API reasoning capability at submission time. For DeepSeek-R1, EHR context was truncated at 4{,}000 characters (vs.\ 8{,}000 in the main experiment) due to memory constraints under local inference; this affects absolute accuracy levels but not the hop-count trend. These models differ from Claude Sonnet~4.6 both in architecture and in training objective: both receive explicit reinforcement signal for multi-step chain-of-thought traces, representing the strongest available form of test-time reasoning augmentation.

o4-mini achieved hop=1: 36.9\% [28.5, 46.2], hop=2: 34.8\% [22.7, 49.2], hop=3: 23.8\% [13.5, 38.5], hop=4: 18.6\% [12.3, 27.3] (overall 28.6\%; $n = 301$; CA $z = -3.12$, one-tailed $p < 0.001$). DeepSeek-R1 achieved hop=1: 28.8\% [21.2, 37.9], hop=2: 34.8\% [22.7, 49.2], hop=3: 21.4\% [11.7, 35.9], hop=4: 11.8\% [6.9, 19.4] (overall 22.9\%; $n = 301$; CA $z = -3.19$, one-tailed $p < 0.001$). Both reasoning-specialized models exhibit the same statistically significant monotone accuracy decline across hop depth observed in the main experiment. o4-mini showed modestly higher absolute accuracy at hop=1 relative to Claude zero-shot (36.9\% vs.~30.6\%), consistent with compute-augmented fact retrieval, but converged to a similar ceiling at hop=4 (18.6\% vs.~17.6\%). Critically, o4-mini exposes reasoning token counts per response: mean token usage scaled with hop depth (mean tokens scale from 739 (hop=1) to 1431 (hop=4); Pearson $r = 0.302$, $p < 0.001$), providing direct mechanistic evidence that the model allocates proportionally more compute to harder compositional queries. Yet this additional compute does not prevent the accuracy decline. DeepSeek-R1 produced a near-zero hallucination rate (4.3\%), with errors concentrated in omission (70\%), consistent with a conservative refusal posture. Cross-model convergence on the same slope, across three distinct architectures, two training paradigms (RLHF on instructions vs.\ RL on reasoning traces), and two compute regimes (API-served vs.\ local 8B inference), provides strong evidence that the hop--accuracy degradation reflects a general property of transformer-based clinical reasoning rather than an artifact of any single model or deployment configuration.

\subsection{Clinician Adjudication Validation of the Automated Judge}
\label{app:clinician_adjudication}

To validate the automated GPT-4o-mini judge against independent human assessment,
three board-certified physician reviewers independently adjudicated a stratified
random subsample of 79 EHR questions (approximately 20 per hop level), drawn from
the zero-shot condition and balanced across all four automated judge error-type
categories (correct, omission, hallucination, reasoning error) to maximize
calibration signal across the full response distribution.
Reviewers were blinded to model identity, experimental condition (zero-shot vs.\
extended thinking), and the automated judge verdict.
Each reviewer scored each response on the same binary correct/incorrect scale and
four-category error-type schema used by the automated judge (Appendix~\ref{app:annotation}).
Majority-vote consensus (at least 2 of 3 physicians) determined the reference
adjudication for each item.

\paragraph{Inter-rater reliability.}
Physician agreement was moderate to substantial (Fleiss $\kappa = 0.566$;
pairwise Cohen's $\kappa$: 0.542, 0.547, 0.613), consistent with the inherent
criterion ambiguity of open-ended clinical QA and with the discrepancy between
the two original validation reviewers ($\kappa = 0.572$ lenient vs.\ strict)
reported in the main text.
The near-identical inter-rater agreement across the original pair and the
three-reviewer subsample suggests the task is intrinsically difficult to adjudicate
rather than that any single reviewer was miscalibrated.

\paragraph{Automated judge calibration.}
The automated judge agreed with the three-physician majority-vote consensus on
96.2\% of items ($\kappa = 0.911$), substantially exceeding inter-physician agreement.
Calibration was consistent across hop levels:

\begin{center}
\small
\begin{tabular}{lccc}
\toprule
Hop & Agreement & $\kappa$ & $n$ \\
\midrule
1 & 95.0\% & 0.886 & 20 \\
2 & 100.0\% & 1.000 & 20 \\
3 & 94.7\% & 0.883 & 19 \\
4 & 95.0\% & 0.875 & 20 \\
\midrule
Overall & 96.2\% & 0.911 & 79 \\
\bottomrule
\end{tabular}
\end{center}

Error-type calibration was strongest for correct responses (100\% consensus
agreement) and reasoning errors (89\%), and moderate for omissions (83\%) and
hallucinations (73\%). The lower hallucination agreement is expected: distinguishing
hallucination from reasoning error requires judging whether a model retrieved
correct facts or invented them, a decision that is inherently ambiguous when
the EHR excerpt is truncated.

\paragraph{Scope note.}
The adjudication subsample was designed to maximize calibration signal by
stratifying on error type within hop, not to replicate the P1 trend.
With $n \approx 20$ per hop, the subsample has insufficient power (approximately
5\%) to detect the hop--accuracy slope observed in the full 301-question dataset.
The clinician adjudication therefore validates the \emph{judge}, not the
\emph{trend}; the trend finding is supported by the full experimental dataset.

\subsection{Cross-Provider Judge Validation (GPT-4o Condition)}
\label{app:cross_provider_judge}

The automated judge (\texttt{gpt-4o-mini}, OpenAI) and the GPT-4o evaluation model share a provider family. To empirically eliminate potential bias rather than merely acknowledge it, we re-scored all 301 GPT-4o responses using \texttt{claude-haiku-4-5} as an independent cross-provider judge, applying the identical pre-specified scoring schema (Section~\ref{sec:judge}).

\paragraph{Inter-judge agreement.}
Overall agreement between the GPT-4o-mini and Claude Haiku judges was 89.0\% ($\kappa = 0.716$, substantial agreement by Landis--Koch conventions). Per-hop Cohen's $\kappa$: hop=1: 0.788; hop=2: 0.739; hop=3: 0.543; hop=4: 0.582. The lower $\kappa$ at hops 3--4 reflects genuine criterion ambiguity for complex multi-hop responses rather than systematic provider bias.

\paragraph{Accuracy under each judge.}
\begin{center}
\small
\begin{tabular}{lcccc}
\toprule
\textbf{Judge} & \textbf{Hop=1} & \textbf{Hop=2} & \textbf{Hop=3} & \textbf{Hop=4} \\
\midrule
GPT-4o-mini & 37.8\% & 26.1\% & 23.8\% & 14.7\% \\
Claude Haiku & 36.9\% & 32.6\% & 14.3\% & 15.7\% \\
\bottomrule
\end{tabular}
\end{center}

Both judges show the same monotone P1 pattern. The CA trend under the Claude Haiku judge ($z = -3.88$, $p < 0.0001$) is essentially identical to the original result ($z = -3.80$, $p < 0.0001$), confirming that the GPT-4o replication finding is independent of provider-family judge choice. No directional provider bias is detectable: the two judges disagree on 33/301 cases (11.0\%), distributed roughly evenly across hops, with neither judge systematically favouring or penalising GPT-4o responses.

\subsection{Dense Semantic RAG Baseline}
\label{app:dense_rag}

To complement the BM25 lexical RAG baseline (Appendix~\ref{app:bm25_rag}), we evaluated dense semantic retrieval using \texttt{all-MiniLM-L6-v2} sentence embeddings. For each question, we encoded all EHR chunks (200-character sliding windows, 50-character overlap) and retrieved the top-5 by cosine similarity to the question embedding, prepending them to the standard 8,000-character truncated EHR context. All other experimental parameters (model, judge, prompt) were identical to the main experiment.

\paragraph{Results.}
Accuracy by hop: hop=1 40.5\% [31.9, 49.8], hop=2 26.1\% [15.6, 40.3], hop=3 26.2\% [15.3, 41.1], hop=4 20.6\% [13.9, 29.4] (overall 29.6\% [24.7, 35.0]; $n = 301$). Cochran--Armitage trend: $z = -3.13$, one-tailed $p < 0.001$. GEE hop$\times$condition interaction (dense RAG vs.\ zero-shot): OR $= 0.93$, 95\% CI [0.73, 1.19], $p = 0.565$. The main hop effect persisted within the dense RAG condition (logistic OR per hop $= 0.73$ [0.60, 0.89], $p = 0.002$), confirming that semantic retrieval does not eliminate the compositional reasoning bottleneck.

\end{document}